\documentclass[sigconf]{acmart}

\AtBeginDocument{%
  \providecommand\BibTeX{{%
    \normalfont B\kern-0.5em{\scshape i\kern-0.25em b}\kern-0.8em\TeX}}}

\setcopyright{acmcopyright}
\copyrightyear{2025}
\acmYear{2025}
\acmDOI{XXXXXXX.XXXXXXX}

\acmConference[MM '25]{33rd ACM International Conference on Multimedia}{10.27-10.31, 2025}{Dublin}
\acmPrice{15.00}
\acmISBN{978-1-4503-XXXX-X/25/10}

\usepackage{graphicx}
\usepackage{multirow}
\usepackage{enumitem}
\usepackage{bbding}
\usepackage[capitalize]{cleveref}
\crefname{section}{Sec.}{Secs.}
\Crefname{section}{Section}{Sections}
\Crefname{table}{Table}{Tables}
\crefname{table}{Tab.}{Tabs.}

\usepackage{mathrsfs}
\usepackage{booktabs}
\usepackage{multirow}
\usepackage{multicol}
\usepackage{color}
\usepackage{hyperref}
\usepackage{float}
\usepackage{subfig}
\usepackage{colortbl}
\usepackage{rotating}

\definecolor{LightCyan}{rgb}{0.88,1,1}
\definecolor{iccvblue}{rgb}{0.21,0.49,0.74}
\definecolor{sgreen}{RGB}{30, 150, 30}

\begin{document}

\title{HKD4VLM: A Progressive Hybrid Knowledge Distillation Framework for Robust Multimodal Hallucination and Factuality Detection in VLMs}

\author{Zijian Zhang*}
\email{zhangzijian14@meituan.com}
\affiliation{%
  \institution{Meituan-M17}
  \city{Shanghai}
  \country{China}
}

\author{Xuecheng Wu*$\dagger$}
\email{wuxc3@stu.xjtu.edu.cn}
\affiliation{%
  \institution{Xi'an Jiaotong University}
  \city{Xi'an}
  \country{China}
}

\author{Danlei Huang}
\email{forsummer@stu.xjtu.edu.cn}
\affiliation{%
  \institution{Xi'an Jiaotong University}
  \city{Xi'an}
  \country{China}
}

\author{Siyu Yan}
\email{yansiyu@stu.ecnu.edu.cn}
\affiliation{%
  \institution{East China Normal University}
  \city{Shanghai}
  \country{China}
}

\author{Chong Peng$\ddagger$}
\email{pengchong@meituan.com}
\affiliation{%
  \institution{Meituan-M17}
  \city{Shanghai}
  \country{China}
}

\author{Xuezhi Cao}
\email{caoxuezhi@meituan.com}
\affiliation{%
  \institution{Meituan-M17}
  \city{Shanghai}
  \country{China}
}

\renewcommand{\shortauthors}{Zhang et al.}

\begin{abstract}
Driven by the rapid progress in vision-language models (VLMs), the responsible behavior of large-scale multimodal models has become a prominent research area, particularly focusing on hallucination detection and factuality checking. In this paper, we present the solution for the two tracks of Responsible AI challenge. Inspirations from the general domain demonstrate that a smaller distilled VLM can often outperform a larger VLM that is directly tuned on downstream tasks, while achieving higher efficiency. We thus jointly tackle two tasks from the perspective of knowledge distillation and propose a progressive hybrid knowledge distillation framework termed HKD4VLM. Specifically, the overall framework can be decomposed into Pyramid-like Progressive Online Distillation and Ternary-Coupled Refinement Distillation, hierarchically moving from coarse-grained knowledge alignment to fine-grained refinement. Besides, we further introduce the mapping shift-enhanced inference and diverse augmentation strategies to enhance model performance and robustness. Extensive experimental results demonstrate the effectiveness of our HKD4VLM. Ablation studies provide insights into the critical design choices driving performance gains.

\renewcommand{\thefootnote}{\fnsymbol{footnote}}
\footnotetext[1]{Equal Contribution.}
\footnotetext[2]{Work done during research internship at Meituan-M17.}
\footnotetext[3]{Corresponding author.}
\end{abstract}



\maketitle

\section{Introduction}
\label{sec:intro}

Recently, Large Language Models (LLMs) have made breakthrough advancements~\cite{LLM_review-1,LLM_review-2,LLM_review-3}, demonstrating exceptional capabilities across multiple domains. Inspired by the multimodal cognitive mechanisms of humans, researchers have integrated various visual encoders with the powerful LLM backbones to construct large Vision-Language Models (VLMs)~\cite{MLLM_review-1,TokenFocus-VQA,JTD-UAV,ViC-Bench}. While these models achieve impressive progress in multimodal content understanding, they have also introduced significant challenges for generative AI in real-world applications. Among these, the inherent problem of hallucinations and the challenging issue of factuality checking severely constrain the reliability and safety of multimodal AI systems.

\begin{figure}[t!]
\centering
\includegraphics[width=\linewidth]{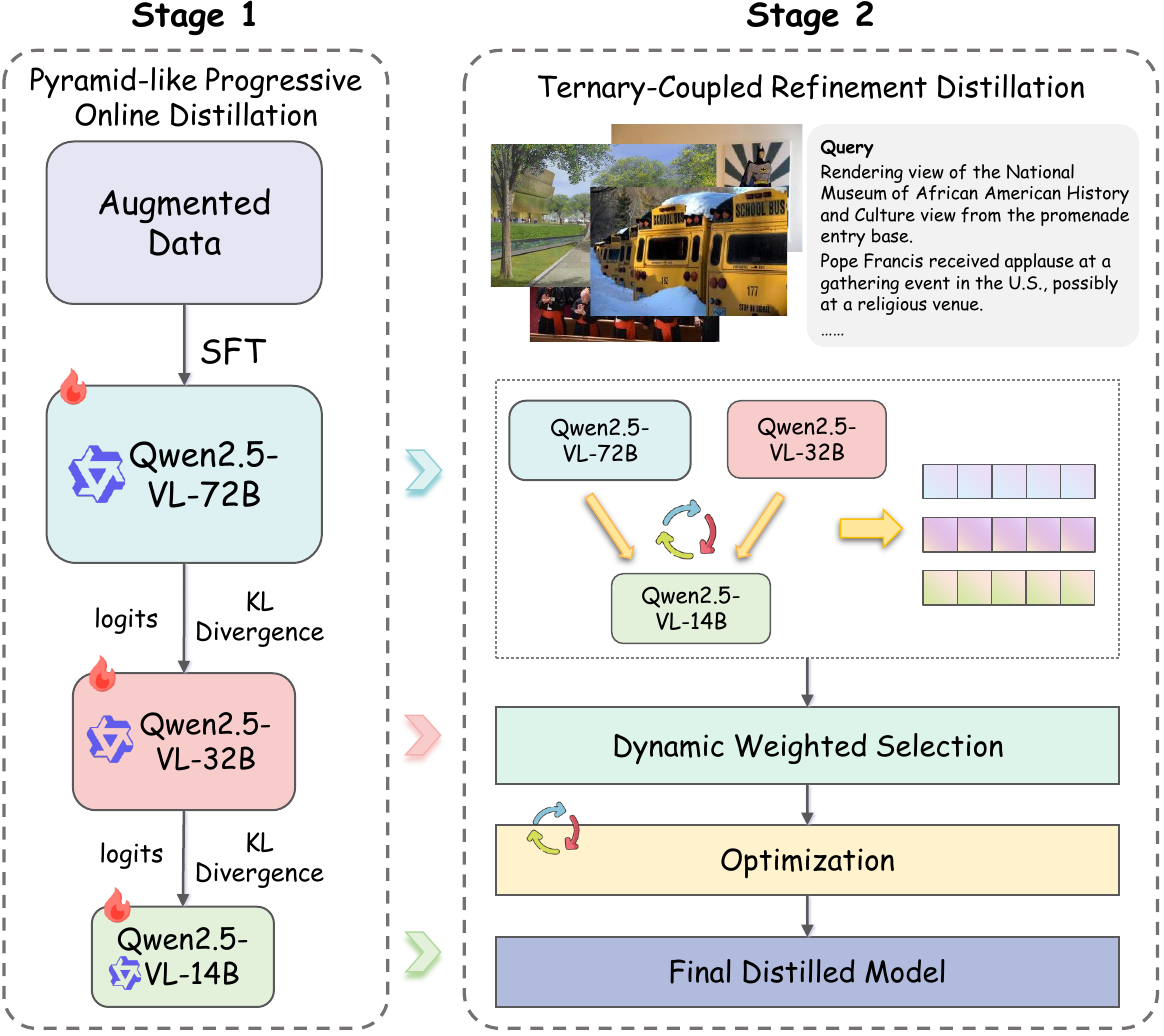}
\vspace{-1.7em}
\caption{The overall illustrations of our proposed HKD4VLM framework, where our core design is the two-stage hybrid knowledge distillation.}
\vspace{-1.7em}
\label{fig:pipelines}
\end{figure}

In light of this, the Responsible AI Challenge explicitly crystallizes the core issues into two tasks: \textbf{(1)}~Multimodal Hallucination Detection, and \textbf{(2)}~Multimodal Factuality Checking, \textit{i.e.},
\begin{itemize}[leftmargin=*]
\item \textbf{Multimodal Hallucination Detection.} This task primarily focuses on identifying hallucinatory content within AI-generated text that is inconsistent with the provided visual input. Detecting such hallucinations is crucial for ensuring the fidelity of generated content in real-world applications.

\item \textbf{Multimodal Factuality Checking.} This task places a greater emphasis on jointly verifying the factuality of a given textual claim using visual-language information, which is designed to address misinformation and biases, representing a critical step in curbing the spread of multimodal disinformation.
\end{itemize}

To tackle the challenges posed by this competition, we approach the problem from the perspective of knowledge distillation for VLMs and propose a progressive hybrid knowledge distillation framework, denoted HKD4VLM. We draw inspiration from the key observations in the general domain, \textit{i.e.}, a smaller distilled VLM can often outperform a larger VLM that is directly tuned on a specialized downstream task, while offering greater computational efficiency~\cite{wang2022efficientvlm,zhang2024vlm}. However, existing knowledge distillation methods for VLMs are predominantly offline~\cite{wang2022efficientvlm,zhang2024vlm}, which results in the significant knowledge gap and prevents the full potential of VLM from being fully unleashed. This limitation is particularly pronounced in fine-grained and factually demanding scenarios, such as multimodal hallucination detection and factuality checking. In contrast, online knowledge distillation, with its mechanism of mutual learning and co-evolution among models, offers a promising way to bridge this gap. Inspired by this, the training phase of our HKD4VLM framework (Fig.~\ref{fig:pipelines}) can be decoupled into two stages: \textbf{(1)} Pyramid-like Progressive Online Distillation and \textbf{(2)} Ternary-Coupled Refinement Distillation. This design enables a hierarchical and progressive learning process, moving from coarse-grained knowledge alignment to fine-grained refinement. Besides, considering the strong correlations between two tasks, we further construct a multi-task learning paradigm, which can provide a richer supervisory signal for online distillation, establishing a joint objective that can promote cross-task knowledge transfer and synergistic gains. During inference, we introduce the Mapping Shift-Enhanced Inference strategy to improve robustness against positional biases by strengthening the semantic alignment between option labels and their corresponding content. Cross-validation and data augmentation strategies are also employed to further enhance model performance and generalization. We perform extensive experiments on the official testbeds of two tasks to verify the effectiveness of our proposed method. Remarkably, our HKD4VLM ranks the first palce on the leaderboards for both tasks, outperforming the second-place entry by a significant margin of up to 14\% F1-score.

This work aims to facilitate the development of specialized and advanced methods for robust multimodal hallucination detection and factuality checking, seeking to effectively mitigate the potential risks of multimodal AI systems in practical applications. In summary, the main contributions of this paper are three-fold:

\begin{itemize}[leftmargin=*]

\item~We pioneer the introduction of knowledge distillation to jointly tackle multimodal hallucination detection and factuality checking in VLMs, thereby establishing a new promising avenue for developing more responsible AI systems.

\item~We propose a novel progressive hybrid knowledge distillation framework (\textit{i.e.}, HKD4VLM), which features a hierarchical two-stage training process. Furthermore, we introduce mapping shift-enhanced inference and diverse data utilization strategies to improve performance and robustness.

\item~Extensive experimental results demonstrate the effectiveness of our HKD4VLM, achieving state-of-the-art performance on two testbeds. We also explore the impact factors driving performance enhancements by a series of ablation studies.

\end{itemize}

\section{Related Work}
\label{sec:Related Work}

\subsection{Vision-Language Models}
With the rapid advancement of large language models~\cite{glm2024chatglm,Llama3_touvron2023llama}, Vision-Language Models (VLMs)~\cite{MLLM_review-3,Qwen2-vl} have emerged, integrating visual perception and textual reasoning to tackle tasks such as image captioning, visual question answering, and factuality understanding. VLMs can be primarily categorized into three main architectures based on their multimodal integration strategies: \textbf{(1)} Feature mapping with concatenation, as seen in PaLM-E~\cite{30_driess2023palm}, LLaVA~\cite{29_liu2024visual}, and CogVLM~\cite{32_wang2023cogvlm}, where visual features are mapped into the LLM’s representation space using MLPs and concatenated with text, offering efficiency but limited interaction between modalities; \textbf{(2)} Query-based cross-attention mechanisms, used by InstructBLIP~\cite{4_dai2023instructblip}, Mini-GPT4~\cite{zhu2023minigpt}, and Qwen-VL~\cite{28_bai2023qwen}, which dynamically fuse cross-modal data via cross-attention for deeper interactions at the cost of higher computational overhead; and \textbf{(3)} Cross-attention within the LLM architecture, as employed by Flamingo~\cite{35_alayrac2022flamingo} and IDEFICS~\cite{37_laurenccon2024obelics}, where visual features are fused directly into intermediate textual representations for seamless alignment. Despite these advancements, ensuring the reliability and factuality of their generated content remains a critical challenge. When tackling above issues, current approaches often exhibit significant limitations. For example, \cite{wang2022efficientvlm,zhang2024vlm} mainly rely on the supervised fine-tuning, which is unsatisfactory and inefficient. Similarly, retrieval-augmented generation (RAG) for factuality checking is frequently bottlenecked by the reliability of the retrieval module~\cite{RAG_1,RAG_2}, which can introduce noise or outdated information, highlighting a critical need for a more efficient and holistic paradigm.

\subsection{Hallucination and Factuality Detection}
Since LLMs often generate fluent and convincing but factually inaccurate outputs, the hallucination issue becomes more challenging to address~\cite{ji2023survey}. Hallucination detection aims to verify whether the generated output is consistent with the inputs. Falke et al.~\cite{falke2019ranking} deploy a classifier to identify the contradictions between LLM output and contexts. Recently, VLMs have made breakthroughs in the cross-modal field, but multimodal hallucinations~\cite{RAG_1,MLLM_review-1} still lead to responses misaligned with visual input. Discriminative approaches for hallucination detection typically adopt the question-answer paradigm~\cite{lovenia2023negative}. While generative approaches often expand the hallucination scope to cover a broader detection range, such as POPE~\cite{li2023evaluating} and robust instruction tuning~\cite{liu2023mitigating}.

Factuality checking aims to determine whether a statement can reflect the real-world facts. Early research predominantly focus on textual approaches, leveraging linguistic features~\cite{saikh2019deep}, knowledge graphs~\cite{liu2020graph}, and supervised classifiers~\cite{yang2018classifier} to determine the veracity of claims. With the emergence of LLMs, factuality detection greatly benefits from advanced contextual understanding to learn complex relationships between claims and supporting documents~\cite{liu2024llm,tang2024minicheck}. However, many real-world scenarios involve multimodal data, where claims typically include both textual and visual information, resulting in the increasing interest in factuality checking for VLMs. Recent works~\cite{FC_1,FC_2} primarily extend LLMs by enabling evidence retrieval, grounding textual claims in images, and training models to discern factual consistency across modalities.

\subsection{Knowledge Distillation in VLMs}
Knowledge distillation is a kind of popular technique where a student model is trained to imitate the outputs of a powerful teacher model, making model compression possible with little loss in downstream performance~\cite{hinton2015distilling}. In recent years, this technique has been widely applied in VLMs to enhance efficiency and scalability in multi-modal tasks. Early works~\cite{29_liu2024visual,zhu2023minigpt} mainly adopt offline distillation, where the student model learns from predetermined teacher outputs. However, this static approach often leads to a knowledge gap because there is no mutual adaptation between teacher and student during training. To address this issue, researchers have started to investigate online distillation~\cite{zhang2017deepmutuallearning}, in which teacher and student models are trained together, enabling mutual adaptation and more effective knowledge transfer. Despite its promise, online distillation is still rarely studied in the VLM domain. Notably, Yu et al.~\cite{yu2024selectdistill} propose a dual-teacher online distillation framework, allowing students to adaptively learn from both fine-tuned and pretrained teacher models,which helps to prevent catastrophic forgetting and maintain zero-shot learning ability in continual vision-language learning. Building on these advances, our work introduces the progressive hybrid online distillation combined with multi-task learning, intending to develop more accurate and robust VLMs for the multimodal verification scenarios.
\section{Methodology}
\label{sec:Mthods}

\subsection{Preliminaries}
\label{3.1}

\noindent \textbf{Qwen-2.5-VL Model Family.}~Currently, Qwen-2.5-VL~\cite{Qwen2d5-vl} series represents a family of state-of-the-art VLMs. Its core design comprises a novel ViT encoder, a backend language model , and the MLP-based connector. Qwen-2.5-VL optimizes its vision encoder by introducing native dynamic resolution processing and window attention, enabling efficient handling of images and videos while preserving original fidelity. The connector effectively integrates visual features into the backend LLM by grouping spatially adjacent patch features and projecting them via MLPs. To meet the requirements of diverse application scenarios, Qwen-2.5-VL series consist of varying capacities, ranging from 3B to 72B.

\begin{figure}[t!]
\centering
\includegraphics[width=\linewidth]{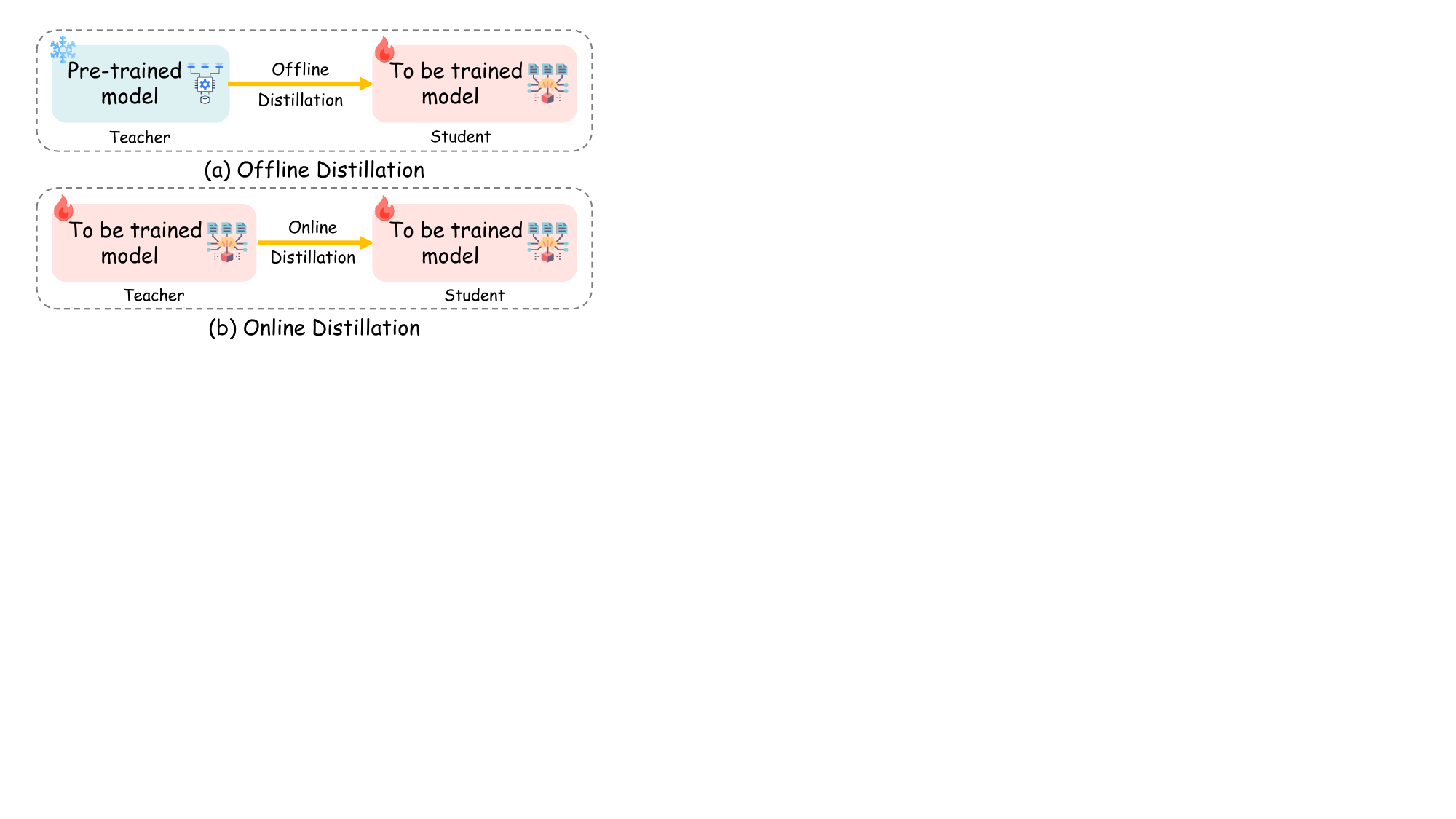}
\vspace{-2.0em}
\caption{The illustrations of online knowledge distillation.}
\vspace{-1.45em}
\label{fig:OD}
\end{figure}

\noindent \textbf{Online Knowledge Distillation.}~Unlike offline distillation methods, which require a powerful pre-trained teacher model, online knowledge distillation employs a collaborative learning paradigm where multiple peer models are trained simultaneously, learning from each other and evolving together, as illustrated in Fig.~\ref{fig:OD}. This approach circumvents the challenges of selecting an appropriate teacher model and the inefficiency of a cascaded training process inherent in offline distillation approaches. In the online distillation framework, there is no fixed teacher. Instead, at each training step, one model can dynamically act as a "teacher" for its peers. For instance, in a cohort of $N$ student models $\{S_1, S_2, \dots, S_N\}$, the total loss function for any given student $S_i$ can be formulated as:
\begin{equation}
\mathcal{L}_{S_i} = \mathcal{L}_{\mathbf{CE}}(y, \sigma(z_i)) + \beta \cdot \sum_{j=1, j\neq i}^{N} \mathcal{L}_{\mathbf{Distill}}(z_j, z_i),
\end{equation}
where $z_i$ represents the predictions output by the student model $S_i$. $\mathcal{L}_{\mathbf{CE}}$ is the standard cross-entropy loss against the ground-truth labels $y$. $\mathcal{L}_{\mathbf{Distill}}$ is the mutual distillation loss between student $S_i$ and its peer $S_j$. Besides, $\beta$ is the hyperparameter that can balance the strengths of mutual learning component.

\subsection{Two-Stage Hybrid Knowledge Distillation}
\label{3.3}
\noindent \textbf{Pyramid-like Progressive Online Distillation.}~As displayed in Fig.~\ref{fig:pipelines}, our primary objective is to establish a coarse-grained alignment between base models and the domain-specific knowledge. Specifically, we select 72B, 32B, and 14B versions of Qwen-2.5-VL~\cite{Qwen2d5-vl} to respectively serve as large, medium, and small base models. We first perform cold-start using supervised fine-tuning for Qwen-2.5-VL-72B on the combined official dataset (as detailed in Sec.~\ref{3.5}), allowing models to fully adapt to the distributions and patterns of downstream tasks. After that, a conventional offline distillation approach would treat the 72B model as the teacher and the 32B model as the student, training the latter to mimic the former's behavior. However, our empirical experimental results demonstrate that this method yields suboptimal performance on our coupled tasks. We attribute this to the inherent knowledge gap exists between the teacher and student models. Moreover, this approach fails to fully unleash the student's comprehension potential in fine-grained and factually demanding scenarios. As a result, we select the online distillation, which features a teacher-student co-evolution mechanism. Subsequently, considering that knowledge is primarily transferred between models by matching their output logits, we finally choose the most representative response-based paradigm, which relies on directly distilling knowledge from these logits.

Specifically, the raw and unnormalized output vectors Logits ($z$), which come from the model's final layer before the function of Softmax, encode not only the most likely predictions but also richer information about the relationships between different classes, leading to the emergence of dark knowledge, which is crucial for improving generalization of VLMs. To effectively transfer this knowledge, the Kullback-Leibler (KL) divergence with a temperature coefficient $\tau$ is employed as the distillation loss. As a result, the overall process can be formulated as:
\begin{equation}
\mathcal{L}_{\mathbf{Distill}}(z_j, z_i) = \tau^2 \cdot D_{\mathbf{KL}}(\sigma(z_j / \tau) || \sigma(z_i / \tau)),
\end{equation}
where $\tau$ is deployed to smooth the probability distributions derived from the logits. A higher temperature could soften the logits outputs, amplifying the probability values of less likely classes, which enables student model to learn more detailed inter-class structural information from its peer teacher, rather than merely mimicking its most confident prediction. $\tau^2$ is targetedly included to properly scale the gradients with respect to the temperature, which facilitates to stabilize the overall training process.

Through our meticulous online distillation paradigm, we first co-train the medium model with large model in a shared high-dimensional latent space. This initial stage allows the medium model to absorb the complex discriminative knowledge from its larger counterpart, thereby enhancing its own representation capability in high-dimensional semantics. Subsequently, we conduct the second stage of online distillation between the distillated medium model and targeted small model, which can effectively transfer the refined discriminative knowledge to the most compact model, significantly boosting its performance ceiling while preserving its lightweight advantages. Through above efforts, we ultimately establish a pyramid-like progressive online distillation framework at the capacity-aware level, which efficiently achieves the cascaded transfer of knowledge from the large model to the small model, \textit{i.e.},
\begin{align}
\mathcal{L}_{\mathbf{overall}} &= \left( \mathcal{L}_{\mathbf{CE}}(y, z_{S_M}) + \alpha \mathcal{L}_{\mathbf{KD}}(z_{S_L}, z_{S_M}) \right) \nonumber \\
                             &\quad + \left( \mathcal{L}_{\mathbf{CE}}(y, z_{S_S}) + \beta \mathcal{L}_{\mathbf{KD}}(z_{S_M}, z_{S_S}) \right),
\end{align}
where $\mathcal{L}_{\mathbf{KD}}$ is the Kullback-Leibler divergence-based distillation loss between logits of teacher-student pairs. $\alpha$ and $\beta$ are hyperparameters that balance the weight of distillation losses for each stage. $S_L$ is the large model, which acts as the initial teacher for medium and small models (\textit{i.e.}, $S_M$ \& $S_S$).

\begin{figure}[t!]
\centering
\includegraphics[width=\linewidth]{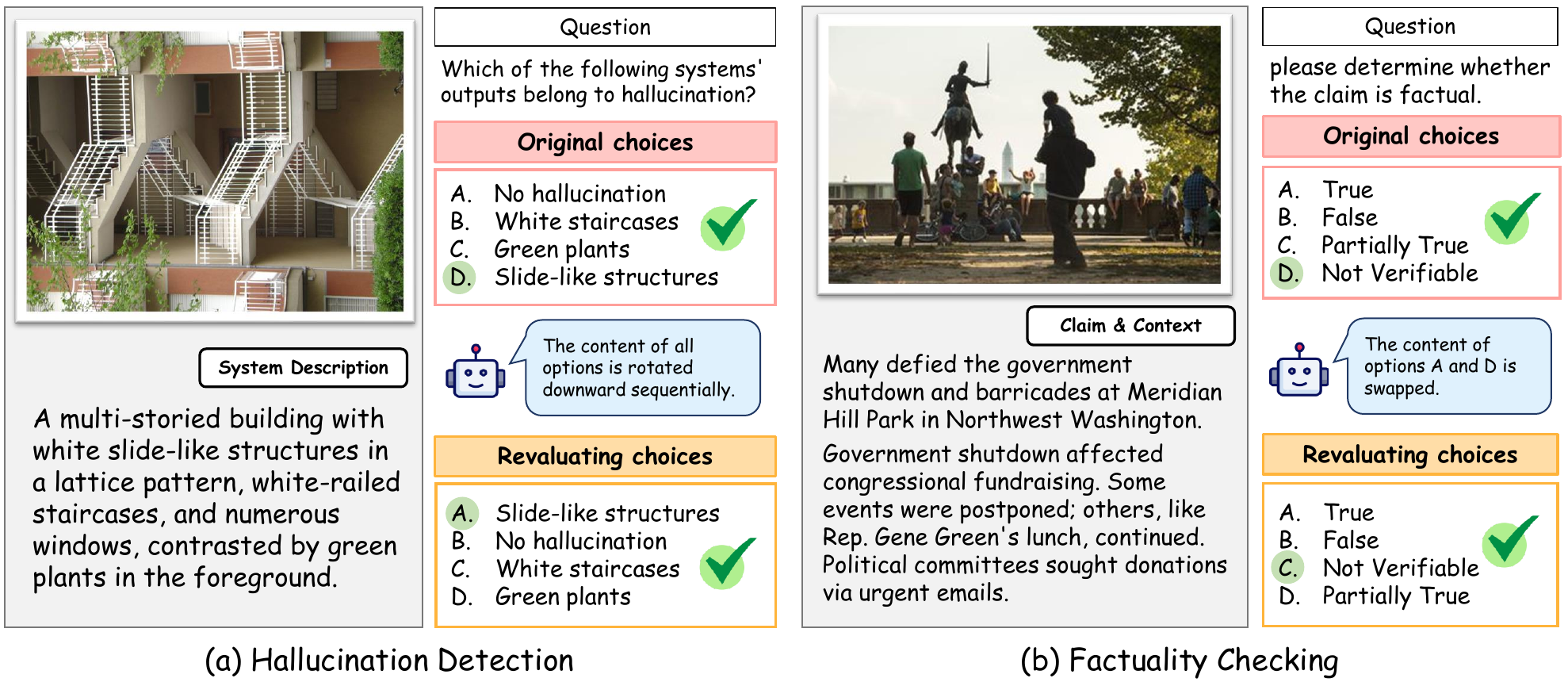}
\vspace{-2.0em}
\caption{The overall illustrations of our proposed mapping shift-enhanced inference strategy.}
\vspace{-1.4em}
\label{fig:MSEI}
\end{figure}

\noindent \textbf{Ternary-Coupled Refinement Distillation.}~To further enhance the synergistic learning dynamics and unlock a more profound understanding of downstream data distribution, we introduce a supplementary refinement stage. Inspired by the distillation strategy in Intern-VL series VLMs~\cite{Internvl,InternVL3}, we propose the Ternary-Coupled Refinement Distillation (TCRD) stage. Concretely, the large, medium, and small models are further coupled to learn collaboratively. All three models process the same images and prompts, and the overall loss is computed via a joint KL divergence across their output logits, forcing students to learn from the teacher in the multi-student and competitive environment. Besides, the ternary distillation loss $\mathcal{L}_{\mathbf{TernaryKD}}$ is dynamically weighted, and the training objective is to simultaneously distill knowledge from both the large and medium models to the small model, with a controlled focus, \textit{i.e.}, 
\begin{equation}
\mathcal{L}_{\mathbf{TernaryKD}} = \gamma \cdot \mathcal{L}_{\mathbf{KD}}(z_{S_L}, z_{S_S}) + (1-\gamma) \cdot \mathcal{L}_{\mathbf{KD}}(z_{S_M}, z_{S_S}),
\end{equation}
where $z$ represents the output logits. The hyperparameter $\gamma \in [0, 1]$ serves as a dynamic weighting factor. A larger value of $\gamma$ places greater emphasis on refining the small model by pushing it to more closely mimic the output distribution of large model. Conversely, a smaller $\gamma$ prioritizes the distillation from the small model to the medium model. This dynamic weighting allows us to flexibly control the knowledge flow and adjust the distillation process based on the specific refinement goals at each stage.

\subsection{Mapping Shift-Enhanced Inference}
\label{3.4}
To enhance model robustness and mitigate \textit{shortcut learning} on the targeted tasks under multiple-choice QA settings, we introduce the Mapping Shift-Enhanced Inference strategy, as illustrated in Fig.~\ref{fig:MSEI} above. This method can act as the \textit{cognitive shell game} to ensure the model comprehends the deep semantic content rather than relying on positional biases. For each question, the model first performs a standard inference to select an answer. Subsequently, we permute the textual content between the option labels (\textit{e.g}., swapping the content of option letters \textit{A} and \textit{B}) and task the model with re-evaluating its choice. This process compels the model to decouple the superficial option letter from its underlying meaning, facilitating it to learn the invariant mapping between user query and the ground-truth content. By successfully tracking the answer through this shuffle, our distilled VLM demonstrates a more profound and robust understanding, proving its capabilities extend beyond exploiting the simple dataset heuristics.

\begin{table}[t!]
\centering
\caption{Statistics of Multimodal Hallucination Detection.}
\vspace{-0.7em}
\setlength{\arrayrulewidth}{0.40pt}
\renewcommand{\arraystretch}{1.0}
\resizebox{\linewidth}{!}{%
\begin{tabular}{lcccc}
\toprule
Field & Train & Dev & Test \\ 
\midrule
Number of images & 2,800 & 200 & 1,000 \\
Average caption length & 40.3 words & 40.1 words & 40.4 words \\
Hallucination types & 3 & 3 & 3 \\
Ground-truth annotations & 3,000 & 250 & 1,500 \\
Scene categories & 18 & 18 & 18 \\
\hline
\end{tabular}
}
\vspace{-1.0em}
\label{tab:statistics-MHD}
\end{table}

\begin{table}[t!]
\centering
\caption{Statistics of Multimodal Factuality Checking.}
\vspace{-0.7em}
\setlength{\arrayrulewidth}{0.40pt}
\renewcommand{\arraystretch}{1.0}
\resizebox{\linewidth}{!}{%
\begin{tabular}{lcccc}
\toprule
Field & Train & Dev & Test \\ 
\midrule
Number of claims & 2,800 & 200 & 1,000 \\
Average context length & 120 words & 115 words & 122 words \\
Number of images & 2,800 & 200 & 1,000 \\
Factuality labels & 4 & 4 & 4 \\
Scene categories & 16 & 16 & 16 \\
\hline
\end{tabular}
}
\vspace{-1.4em}
\label{tab:statistics-MFC}
\end{table}

\begin{table*}[t!]
\centering
\caption{Performace comparisons of state-of-the-art VLMs and our proposed HKD4VLM in terms of Precision, Recall, and F1-Score metrics~(\%) on two tasks. Note that we highlight the best performance in \textbf{\textit{bold}} and \underline{underline} the second performance.}
\vspace{-0.5em}
\setlength{\arrayrulewidth}{0.40pt}
\renewcommand\arraystretch{1.0}
\resizebox{\textwidth}{!}{
\begin{tabular}{lccccccccccccc}
\toprule[0.40pt]
\multirow{2}{*}{Method} & \multirow{2}{*}{Organization}    & \multicolumn{3}{c}{Multi-modal Hallucination Detection} & \multicolumn{3}{c}{Multi-modal Fact Checking} \\
\cmidrule(lr){3-5} \cmidrule(lr){6-8}

&  & Precision  & Recall  & F1-Score  & Precision  & Recall  & F1-Score    \\ 
\midrule
GPT-4o-20240513~\cite{openai2024gpt4o}  & OpenAI      & \underline{72.0}            & 62.0            & \underline{66.7}            & \underline{78.0}         & \underline{66.4}         & \underline{71.7}    \\
Hunyuan-Vision-20250103~\cite{Hunyuan-Vision-20250103} & Tencent  & 47.2            & 54.0            & 50.4            & 49.4         & 55.8         & 52.4    \\
Qwen2.5-VL-7B~\cite{Qwen2d5-vl}   & Alibaba & 37.2            & 33.6            & 35.3           & 40.2         & 39.7         & 39.9        \\
Qwen2.5-VL-14B~\cite{Qwen2d5-vl} & Alibaba  & 48.4            & 54.4            & 51.2            & 50.2         & 56.1         & 53.0        \\
Qwen2.5-VL-32B~\cite{Qwen2d5-vl} & Alibaba  &  48.3    &  51.1    &  49.6         &  50.8              &  52.8               &  51.8             \\
Qwen2.5-VL-72B~\cite{Qwen2d5-vl}  & Alibaba  & 50.6            & 55.8            & 53.1            & 53.6         & 58.9         & 56.1        \\
Qwen-QVQ-72B~\cite{team2024qvq} & Alibaba  & 62.4            & \underline{64.6}            & 63.5            & 57.8         & 62.0         & 59.8        \\
InternVL-2.5-8B~\cite{Internvl} & Shanghai AI Lab    & 38.4            & 36.2            & 37.3            & 41.6         & 44.1         & 42.8        \\
InternVL-2.5-78B~\cite{Internvl} & Shanghai AI Lab    & 53.6            & 59.3            & 56.3            & 54.1         & 60.2         & 57.0        \\

InternVL-3-8B~\cite{InternVL3}  & Shanghai AI Lab  &  39.2     &  37.8    &  38.5         &  44.8              &  48.4               &  46.5              \\
InternVL-3-78B~\cite{InternVL3}  & Shanghai AI Lab &  59.5    &  64.2    &  61.8         &  63.4              &  60.2               &  61.8              \\

\rowcolor{iccvblue!20}
\textbf{HKD4VLM (Ours)} & \textbf{--}   & \textbf{97.0}            & \textbf{99.4}            & \textbf{98.2}           & \textbf{97.8}         & \textbf{99.0}         & \textbf{98.4}    \\
\hline
\end{tabular}
}
\vspace{-0.4em}
\label{tab:main-compare}
\end{table*}

\begin{table}[t!]
\centering
\caption{The overall ablation studies of introduced components in terms of F1-Score (\%) on two tasks.}
\vspace{-0.5em}                   
\setlength{\arrayrulewidth}{0.40pt}
\renewcommand{\arraystretch}{1.0}
\resizebox{\linewidth}{!}{%
\begin{tabular}{lcccc}
\toprule
Method & MHD  & MFC   \\ 
\midrule
Qwen-2.5-VL-14B~\cite{Qwen2d5-vl} (Baseline)          & 51.2 & 53.0 \\
+ Pyramid-like Progressive Online Distillation  & 97.3 (\textcolor{sgreen}{\textbf{+46.1}}) & 97.5 (\textcolor{sgreen}{\textbf{+44.5}}) \\
+ Ternary-Coupled Distillation  & 97.9 (\textcolor{sgreen}{\textbf{+0.6}}) & 98.0  (\textcolor{sgreen}{\textbf{+0.5}}) \\
+ MSEI Strategy & \textbf{98.2} (\textcolor{sgreen}{\textbf{+0.3}}) & \textbf{98.4} (\textcolor{sgreen}{\textbf{+0.4}}) \\
\hline
\end{tabular}
}
\vspace{-1.3em}
\label{tab-overall-ablation}
\end{table}

\subsection{Data Utilization Strategy}
\label{3.5}
\noindent \textbf{Multi-Task Learning Paradigm.}~Given the intrinsic conceptual link between multimodal hallucination detection and multimodal factuality checking—as both require a deep alignment of visual evidence with textual claims—we thus employ a multi-task learning approach. By training a single model on both tasks simultaneously, we enable it to develop a more holistic and generalizable understanding of visual-semantic consistency. This synergistic training process allows the VLMs to leverage features learned from one task to improve its performance on the other task. Besides, this richer learning environment is also well-suited for the online knowledge distillation , as it can provide diverse and challenging signals for the peer models to teach each other.

\noindent \textbf{Five-Fold Cross-Validation.}~To ensure our evaluation is robust and our models can generalize well, we utilize the five-fold cross-validation paradigm. The entire training dataset is partitioned into five non-overlapping folds. In each of the five iterations, we train our models on four of the folds and reserve the fifth fold for fair testing. Note that the final performance scores are the averaged outputs of the metrics across all five test folds.

\noindent \textbf{Data Augmentation.}~We propose the Synergistic Negative Synthesis (SNS) trick for the initial supervised fine-tuning of large model (\textit{i.e.,} Qwen-2.5-VL-72B~\cite{Qwen2d5-vl}) during our pyramid-like distillation. SNS can construct highly realistic negative samples by extracting the specific hallucinated entities in MHD and feeding them into the textual descriptions of the factually correct samples from MFC. This process can generate verisimilar examples containing a single targeted falsehood, compelling the model to develop fine-grained verification skills beyond coarse semantic matching~\cite{wang2024building}. 
\begin{table}[t!]
\centering
\caption{The ablation comparisons of different weighting factors $\gamma$ for our TCRD stage on two tasks.}
\vspace{-0.5em}                       
\setlength{\arrayrulewidth}{0.40pt}
\renewcommand{\arraystretch}{1.0}
\resizebox{\linewidth}{!}{%
\begin{tabular}{ccccccccccc}
\toprule
\multirow{2}{*}{\raisebox{-2ex}{\begin{tabular}[c]{@{}c@{}}Weighting \\Factor~$\gamma$\end{tabular}}} & \multicolumn{3}{c}{MHD} & \multicolumn{3}{c}{MFC} \\ 
\cmidrule(l){2-4}   \cmidrule(l){5-7} 
& Precision  & Recall  & F1-Score  & Precision  & Recall  & F1-Score       \\
\midrule
=~0.75  & 96.2            & 98.5            & 97.3            & 97.2         & \textbf{99.2}         & 98.2 \\
=~0.50  & 96.2            & 98.4            & 97.2            & \underline{97.8}        & 98.8         & \underline{98.3} \\
=~0.25  & \underline{96.8}            & \underline{98.8}            & \underline{97.8}            & \textbf{98.0}         & 98.8         & \textbf{98.4} \\
\rowcolor{iccvblue!20}
\textbf{=~0.10}  & \textbf{97.0}            & \textbf{99.4}            & \textbf{98.2}            & \underline{97.8}         & \underline{99.0}         & \textbf{98.4} \\
\hline
\end{tabular}
}
\vspace{-1.3em}
\label{tab-weighting-factor}
\end{table}

\begin{figure*}[t!]
\centering
\includegraphics[width=\linewidth]{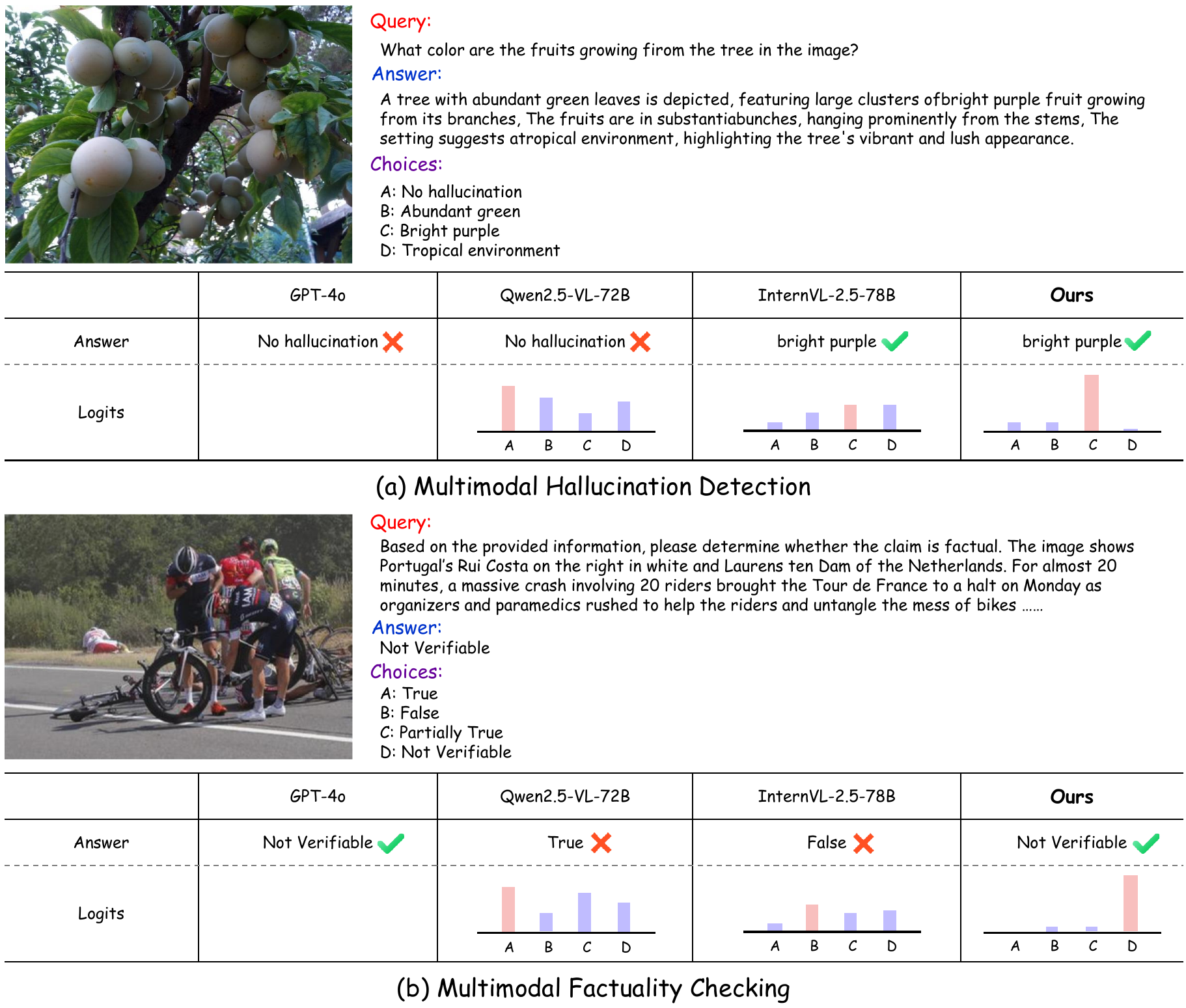}
\vspace{-2.0em}
\caption{Visualization comparisons of HKD4VLM and advanced VLMs in terms of final answer and logits on two tasks.}
\vspace{-0.8em}
\label{fig:case-study}
\end{figure*}

\section{Experiments}
\label{sec:exprs}

\subsection{Implementation Details and Datasets}
\label{4d1}
The overall platform is built upon PyTorch 2.4.1+cu121~\cite{paszke2019pytorch} using a machine with 8 $\times$ NVIDIA A100 GPUs (80G). We empoloy MS-SWIFT (Scalable lightWeight Infrastructure for Fine-Tuning)~\cite{SWIFT} as the base framework. The LoRA~\cite{hu2022lora} is applied for efficient fine-tuning while conducting extensive training on different VLMs of varying capacities. We set the initial learning rate as 1$e$-4 with consine learning schedule, while the other settings keep consistent with the default environment. Besides, we comply with the standardized cross-validation protocol with independent optimization across different splits~\cite{wang2024building,AVF-MAE,wu2023towards}.

The datasets utilized in this challenge are specifically curated for the two distinct tasks. Specifically, Multimodal Hallucination Detection (MHD) task employs a dataset where each image is paired with a targeted caption. Multimodal Factuality Checking (MFC) task uses a dataset where claims are evaluated against both an image and supplementary contextual text. The statistical distributions for both datasets are detailed in Tab.~\ref{tab:statistics-MHD} and Tab.~\ref{tab:statistics-MFC}, respectively. Both datasets are designed to be wide-ranging, featuring a balanced distribution of samples across diverse scenarios, including numerous scene categories (\textit{e.g.}, indoor, outdoor, social, and news), to ensure the robust and generalizable evaluations of our developed models.

\subsection{Evaluation Metrics}
\label{4d2}
To comprehensively evaluate model performance, we deploy various metrics of Precision ($\mathbf{P}$), Recall ($\mathbf{R}$), and F1-score ($\mathbf{F1}$), which are calculated based on the number of True Positives ($\mathbf{TP}$), False Positives ($\mathbf{FP}$), and False Negatives ($\mathbf{FN}$), \textit{i.e.},
\begin{align}
\mathbf{P}   &= \frac{\mathbf{TP}}{\mathbf{TP} + \mathbf{FP}},\\
\mathbf{R}   &= \frac{\mathbf{TP}}{\mathbf{TP} + \mathbf{FN}},\\
\mathbf{F1}  &= 2 \cdot \frac{\mathbf{P} \cdot \mathbf{R}}{\mathbf{P} + \mathbf{R}},
\end{align}
where Precision quantifies the correctness of predictions, and Recall measures the completeness. As the harmonic mean of $\mathbf{P}$ and $\mathbf{R}$, F1-score provides a balanced assessment of performance. Notably, for our joint task, these metrics are computed at the micro-level.

\begin{figure}[t!]
\centering
\includegraphics[width=\linewidth]{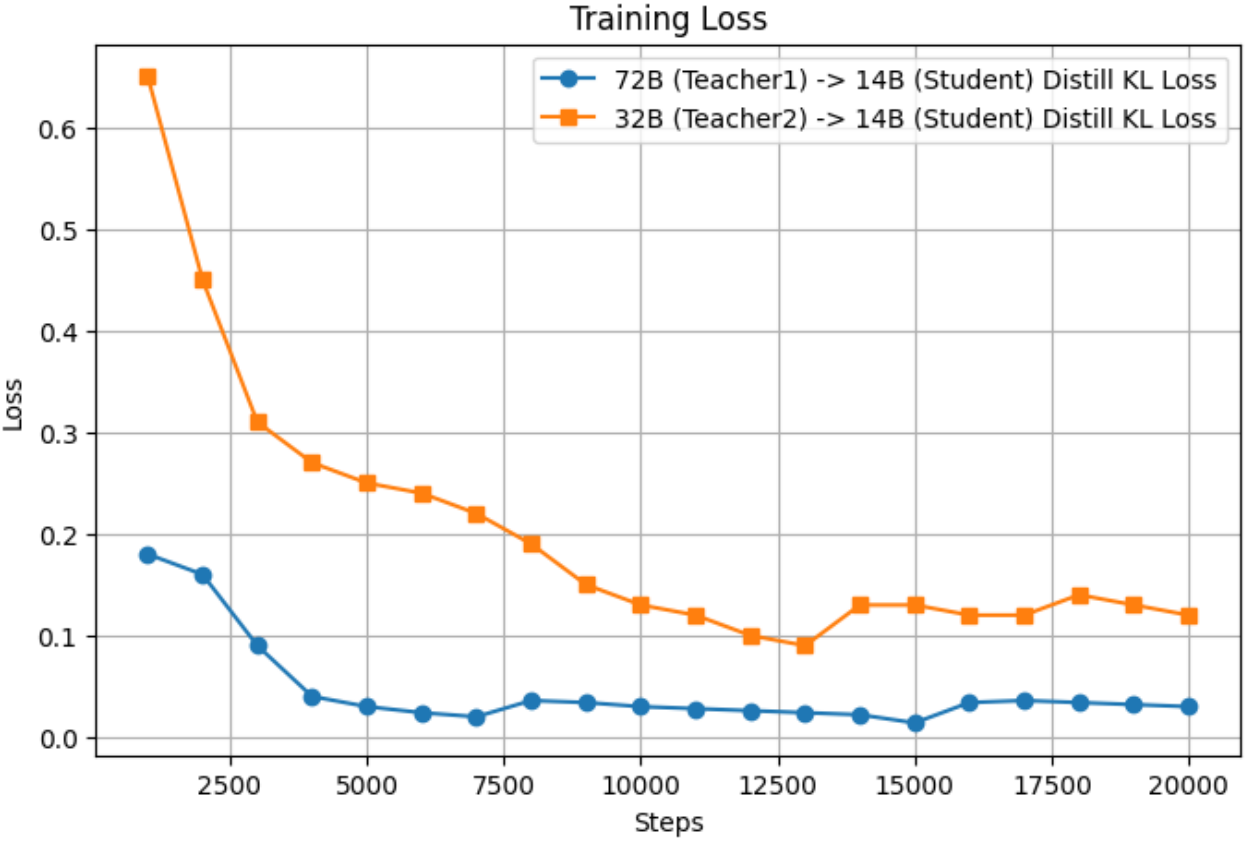}
\vspace{-2.2em}
\caption{The comparisons of loss curves in the ternary-coupled refinement distillation stage.}
\vspace{-1.5em}
\label{fig:Loss_curves}
\end{figure}

\subsection{Performance Comparisons}
\label{4d3}
To validate the efficacy of our proposed framework, we extensively conduct performance comparisons against advanced VLMs, with detailed results presented in Tab.~\ref{tab:main-compare}. These results demonstrate that our HKD4VLM can achieve state-of-the-art performance, significantly outperforming all baselines across both tasks. Specifically, on Multimodal Hallucination Detection (MHD) task, HKD4VLM achieves 98.2\% F1-score, surpassing GPT-4o~\cite{openai2024gpt4o} by a substantial margin of 31.5\%. Meanwhile, on Multimodal Factuality Checking (MFC) task, our approach obtains 98.4\% F1-score, demonstrating a commanding lead over all advanced competitors. Notably, HKD4VLM attains the impressive trade-off between Precision and Recall metrics, indicating its accurate and robust superiority. The above overwhelming performance advantage, achieved with a compact distillated model (\textit{i.e.}, 14B) that outperforms large-scale base model (\textit{e.g.}, Qwen2.5-VL-72B~\cite{Qwen2d5-vl} \& InternVL-3-78B~\cite{InternVL3}), validates the effectiveness of our proposed distillation framework in constructing more responsible and trustworthy multimodal systems.

\subsection{Ablation Study}
\label{4d4}
Tab.~\ref{tab-overall-ablation} displays the results of overall ablation study. With our deverise data utilization strategy, the baseline model achieves modest initial F1-score of 51.2\% on MHD and 53.0\% on MFC. The introduction of Pyramid-like Progressive Online Distillation proves to be the most critical component, yielding a dramatic performance leap of +46.1\% and +44.5\% on two tasks. This substantial gain underscores the fundamental importance of our distillation framework in effectively transferring targeted knowledge from large model to a more compact one. Building upon this strong foundation, our subsequent Ternary-Coupled Refinement Distillation (TCRD) stage provides further refinements, which delivers a consistent enhancement of +0.6\% and +0.5\%. In the end, our MSEI Strategy provides more incremental boost, confirming that enhancing model robustness against input permutations is rather beneficial.

Moreover, we conduct ablation comparisons of weighting factor $\gamma$ in the TCRD stage, as illustrated in Tab.~\ref{tab-weighting-factor} above. Note that MSEI strategy is applied to provide a smooth testbed. From the results, we observe that the overall model achieves better performance when smaller $\gamma$ is assigned, which indicates that the combination of the medium-small model pair can learn more efficiently within the ternary-coupled group. In this setup, the medium model can serve as a better knowledge bridge, providing a higher-quality teacher signal for the final targeted model. Furthermore, as depicted in Fig.~\ref{fig:Loss_curves}, the loss curves for the two distillation pairs in the TCRD stage further corroborate our above conclusions.

\subsection{Case Study}
\label{4d5}
To provide qualitative insights into model performance, we present two cases in Fig.~\ref{fig:case-study}. We observe that HKD4VLM in MHD provides the correct answer with decisive confidence, as evidenced by its sharp logits distribution. In MFC, our approach factually identifies the claim as "Not Verifiable", demonstrating its crucial epistemic capability. This stands in stark contrast to other large models that confidently hallucinate incorrect factual judgments. Overall, both cases indicate that our framework endows VLMs with superior capabilities in both accurate perception and responsible understanding, which is essential for large-scale responsible multimodal AI system.

\section{Conclusions}
\label{sec:Conclusions}
In this paper, we propose the first progressive hybrid knowledge distillation framework termed HKD4VLM for robust multimodal hallucination detection and factuality checking under the multi-task learning paradigm. Our emphatical design consists of the pyramid-like progressive online distillation and ternary-coupled refinement distillation, establishing a comprehensive learning objective that can hierarchically promote cross-task knowledge transfer and synergistic gains. Moreover, we introduce the mapping shift-enhanced inference and diverse data utilization strategies to further boost model performance and robustness. Extensive experimental results impressively demonstrate the effectiveness of our introduced framework. Remarkably, we consistently rank the first place on the offical testbeds across two tasks of Responsible AI challenge.

\clearpage
\newpage
\bibliographystyle{ACM-Reference-Format}
\bibliography{reference}

\end{document}